# Investigating Transfer Learning Capabilities of Vision Transformers and CNNs by Fine-Tuning a Single Trainable Block

Durvesh Malpure* , Onkar Litake* , Rajesh Ingle
SCTR's Pune Institute of Computer Technology
Katraj, Pune
durveshmalpure8@gmail.com, onkarlitake@ieee.org, rbingle@ieee.org

## Abstract

*In recent developments in the field of Computer Vision, a rise is seen in the use of transformer-based architectures. They are surpassing the state-of-the-art set by CNN architectures in accuracy but on the other hand, they are computationally very expensive to train from scratch. As these models are quite recent in the Computer Vision field, there is a need to study it's transfer learning capabilities and compare it with CNNs so that we can understand which architecture is better when applied to real world problems with small data. In this work, we follow a simple yet restrictive method for fine-tuning both CNN and Transformer models pretrained on ImageNet1K on CIFAR-10 and compare them with each other. We only unfreeze the last transformer/encoder or last convolutional block of a model and freeze all the layers before it while adding a simple MLP at the end for classification. This simple modification lets us use the raw learned weights of both these neural networks. From our experiments, we find out that transformers-based architectures not only achieve higher accuracy than CNNs but some transformers even achieve this feat with around 4 times lesser number of parameters.*

## 1. Introduction

Deep Learning based Computer Vision is one of the most researched topics of the last decade[6]. Convolutional Neural Networks introduced by LeCun *et al*. [14] were found to be extremely good at identifying complex features in images. Deep Convolutional Neural Networks that were introduced in 2012 (Krizhevsky *et al*.) in the form of AlexNet[13] with the availability of large datasets such as ImageNet[5] were able to set a new state of the art on the Image classification task. These networks were able to understand complex computer vision tasks such as image classification, semantic segmentation, etc.[16]

*Equal Contribution

Deep Convolutional Neural Networks (CNNs) have seen a number of improvements over the years. Even deeper CNNs were used to improve upon the previous results. Models ranging from VGG-19 from 2015 to EfficientNetV2 from 2021 have continuously improved the state-of-the-art set by CNNs on ImageNet image classification. As CNNs are based on local receptive fields similar to human visual understanding and as they have inductive bias inherent to them, they perform well on the computer vision tasks.[11]

Transformers[23] introduced by Vaswani *et al*. in 2017 have been the default choice of model for all natural language processing tasks as they are the current state-of-the-art surpassing models based on recurrence[24]. These transformer models are based on a major principle of multi head self attention where attention of one word is learned with respect to other words in the sentence and the embeddings are learned with respect to other words. As these models are highly parallelizable unlike Recurrence based networks that process data linearly, they can be trained quickly on multiple GPUs that specialize in parallel computations.

Transformers were first used in computer vision by Dosovitskiy *et al*. where they introduced an architecture called Vision Transformers[7]. These models divide an image into patches and apply multi head self attention (transformer encoder) layers to it. Some improvements to Vision Transformers are Swin Transformers[15] which use hierarchical and shifted windows and calculate attention first locally on a small window and then go to a bigger window size. There have also been some more models introduced that perform better and efficiently. Some examples of such types of models are DeiT[22], CoaT[25] and CaiT[26]. They build on top of vision transformers. VIT-G[27] is the current state of the art on the ImageNet dataset but it has around 2 billion parameters which is far more than the parameters of the ones mentioned before.

Even though there have been a lot of improvements on the image classification task on ImageNet using transformer based architectures, there has not been a proper study of transfer learning capabilities of these new models in a sim-ilar setting. Transfer learning[2] is an important tool which saves the time and resources required for training neural networks for a task from scratch. With the recent shift to these transformer based architectures, we try to investigate how well this family of neural networks perform when transfer learning is applied on the CIFAR-10[12] dataset which has images of size 32*32*3 and are less detailed than the high resolution images in ImageNet. We use openly available ImageNet1K pretrained models from both CNN family and Transformer family of models for Computer Vision and fine-tune them on the CIFAR-10. Fine-tuning can be done using pretrained weights and training the network fully without freezing any layer as well. But we call our fine-tuning restrictive because we only unfreeze the last block of the models to identify which model architecture has the best raw transfer learning capabilities. We do this with some simple changes to the end of the architecture and a training procedure common to all the models to evaluate their capabilities of transfer learning on low resolution images that are upscaled to 224*224*3 with a simple bilinear interpolation. There has not been such a comparison of transfer learning of both transformer based architectures and CNNs on CIFAR-10 to the best of our knowledge.

The main contributions of this paper are:

- Evaluation of CNN and Transformer based models when fine-tuned on CIFAR-10 dataset with a simple yet restrictive fine-tuning.

- A direct comparison between these two families of models with respect to their raw transfer learning capabilities on the basis of their accuracy and efficiency in terms of number of trainable parameters with our setting.

The rest of the paper is structured as follows. Section 2 surveys the advancements in both CNNs and Transformer based models for Vision tasks. Section 3 explains our experimental setup for evaluating different models. Section 4 presents the results we obtain from all the experiments we performed on the models and a discussion on these results. Section 5 presents a conclusion of our paper and some remarks on the future of Transformers for vision related tasks and transfer learning.

## 2. Related Work:

### 2.1. Convolutional Neural Networks

A Convolutional Neural Network (ConvNet/CNN) is a Deep Learning algorithm which takes an image as an input and assigns weights to different layers consisting of filters. A convolution operation is performed between the input and these filters for a forward pass through the model. The learned weights of these filters help to understand features of images automatically. The architecture of a ConvNet is inspired by the organization of the Visual Cortex and is akin to the connectivity pattern of Neurons in the Human Brain. LeNet5 was one of the first and successful convolutional neural networks that were introduced in the field of Deep Learning. It had a fundamental architecture with a sequence of 3 layers: convolution, pooling, and non-linearity.[14]

Krizhevsky *et al*. in 2012 introduced AlexNet[13] which was a deeper and wider version of LeNet. It was able to learn much more complex objects and object hierarchies. It was trained on NVIDIA GTX 580 GPUs which helped reduce its training time and a small revolution was started which stated that large neural networks can be used to solve useful tasks. VGG Network[17] from Oxford was used smaller filters of size 3*3 in each convolutional layer and also combine them as sequences of convolution. It was contrary to the principles of LeNet. Instead of 9*9 or 11*11 filters, they started to become smaller and they were close to becoming 1*1 that LeNet wanted to avoid at least on the first layer of convolution. Szegedy *et al*. [18]., planning to reduce the computational burden of deep neural networks, introduced GoogLeNet, the first Inception architecture. It was a parallel combination of 1*1, 3*3, and 5*5 convolutional filters. It used 1*1 Convolutional blocks to reduce the number of features before expensive parallel blocks. Szegedy *et al*. in February, 2015 introduced Batch Normalization Inception. Later that year in December they released a new version of Inception[19]. He *et al*. in the same month introduced a revolutionary model named ResNet[8]. A residual learning framework was introduced in this model to ease the training of deep neural networks.

Recent developments in the field of Convolutional Neural Network include DenseNet[10] by Huang *et al*. in 2018. In this paper, the observation that convolutional networks can be more accurate and efficient to train if layers close to input and output were shorter was taken into consideration. Each layer was connected to next layer in feed-forward fashion. EfficientNet[20] by Tan *et al*. was introduced in 2020. A new scaling approach was devised that equally scales all depth/width/resolution dimensions using a simple yet highly effective compound coefficient. EfficientNetV2[21] was introduced in the year 2021 which had faster training speed and better efficiency than the previous model. More recently, in 2021, Brock *et al*. introduced NFNets[3]. In this paper, they introduced an adaptive gradient clipping technique and designed Normalizer-Free ResNets. It overcame the undesirable properties of Batch normalization faced due to its dependence on the batch size and interactions between examples. Even though there have many improvements in CNNs, we only consider some classic networks and 2 of the recent ones trained on 224*224 resolution that are in the top 50 leaderboard of the ImageNet

image classification task.[1]

## 2.2. Transformers

Transformers[23] are currently the de facto standard in natural language processing. This encoder-decoder based architecture that relies on multi-head self attention introduced by Vaswani *et al.* in the same paper revolutionized natural language processing. They replaced all the previous architectures for NLP thereafter. Transformers were first used in computer vision by Dosovitskiy *et al.* where they introduced an architecture called Vision Transformers(ViTs)[7]. The Transformer takes in as input, a collection of sub-inputs. A sub-input in NLP is a word. If this word embedding approach was to be blindly followed in computer vision, it would have been computationally very expensive to train a transformer on an image considering every pixel being treated as an individual word leads to quadratic complexity proportional to the square of image size. This size in number of pixels is oftentimes more than hundreds of thousands or even millions of pixels in the case of very high resolution images.[7]

Vision Transformers tackle this problem by dividing every image into small patches and using patch embeddings to pass them through a transformer encoder that calculates self attention[23] for every patch and then at the end passes through fully connected layers to finally classify the image into one of the classes in ImageNet. This approach, while still being moderately computationally expensive and needing a large amount of data to pretrain, could not surpass the state-of-the-art set by CNNs. In December 2020, DeiT (Touvran *et al.*)[22] was proposed as an improvement that scored better than the ViT with a lesser number of parameters and training time. Their teacher-student strategy specific to transformers improved the capability of transformers. A number of transformer based models have been introduced since then such as CoaT[25], Swin Transformers[15], CaiT[26] which have accuracies more than the other in the same order as mentioned. These accuracies reported were top-1 accuracies on ImageNet1k. CoaTs being a model introduced after these two, focus on training a smaller network with very less parameters compared to these two and still achieve more accuracy in comparison with convolutional networks of similar size. While swin transformers use hierarchical windows that calculate attention locally and also in shifted windows which differs from the global attention calculation performed in ViTs, CaiTs build deeper Vision Transformers and achieve even better accuracy compared to Swin Transformers. The current state of the art ViT-G[27] which is a scaled up version of Vision Transformers has 1.843 Billion parameters which is very large compared to the previous models.

## 2.3. Transfer Learning

Although these models are very powerful, they take a long time to train on multi-GPU or TPU systems. One of the largest models, ViT-G, potentially can take more than 10 days on TPUs to get the desired accuracy as mentioned in their report[27]. Not everyone can have access to such a large amount of resources and thus it is impractical to train every model from scratch for every subtask. This is where transfer learning and fine-tuning come into play. Transfer learning uses weights learned from training on a different relevant task or a bigger task of the same domain. When we use a model with pretrained weights, the convergence happens quicker compared to training from scratch. Also if layers of a model are frozen, then the weights of these layers need not be trained further and the training process becomes even faster compared to training the whole model again. We use transfer learning and fine-tune both CNNs and Transformers in our paper to find out which architecture can better transfer it's learned weights and are a better choice for fine-tuning for a specific task.

## 3. Experimental Setup:

We perform our experiments of fine-tuning on CIFAR-10 using models with the setup mentioned below and a general architecture as shown in Figure 1. The code will be made publicly available for reproducibility.

### 3.1. Dataset:

We have selected the CIFAR-10 dataset[12] to fine-tune the models pre-trained only on Imagenet-1K dataset. The ImageNet[5] project is a vast visual database created to aid in the development of visual object recognition software. The Imagenet-1K dataset consists of 1000 classes consisting of 1.2M high resolution samples. The CIFAR-10 dataset contains 60000 32*32 colour images divided into ten classes, each with 6000 images. There are 50000 photos for training and 10,000 images for testing. When passing in as inputs, we upscale the 32*32 images to 224*224 by simple bilinear interpolation. We chose the CIFAR-10 dataset to fine-tune because these images are not very detailed having a resolution of 32*32. We try to find out if the models still perform well with a simple bilinear interpolation to the low resolution images. Another reason for using CIFAR-10 is that, if one were to apply these models to real life vision tasks for special purposes, the amount of data available is usually small. We try to simulate a similar scenario in this experiment.

### 3.2. Models:

We select some purely Convolutional models and some that are purely transformer-based. While selecting, there are a lot of variants of these models we come across. One exam-ple of CNN family could be ResNet101 and ResNet152[8] where the latter is a bigger model. An example of Transformer based models would be ViT-B and ViT-L[7] where the latter is bigger. We select only the largest and best performing models that are openly available as their pretrained versions. As there are a lot of variants of these models, we keep these 2 criterias for the models. The first one being that the model must be trained on a 224*224 resolution and the second one being that the models we use must be pre trained only on ImageNet1k and not ImageNet21K. ImageNet21k includes more classes and images compared to ImageNet1k. No model has seen more data compared to others and has any advantage over the other. This ensures that the comparison of the models is fair and only on the basis of architecture and training procedure. Following this criteria, we select some classic CNN Architectures, some recent CNNs that are comparable to the state of the art and some Transformers comparable to the state of the art. The only exception to these criteria is the ViT-L model which was pretrained on ImageNet21k. We include only a version of this model which is bigger but faster to train as a comparison with the transformers because it was the first model to incorporate transformers into computer vision in an efficient way. We list all the models we take in Table 1 as well.

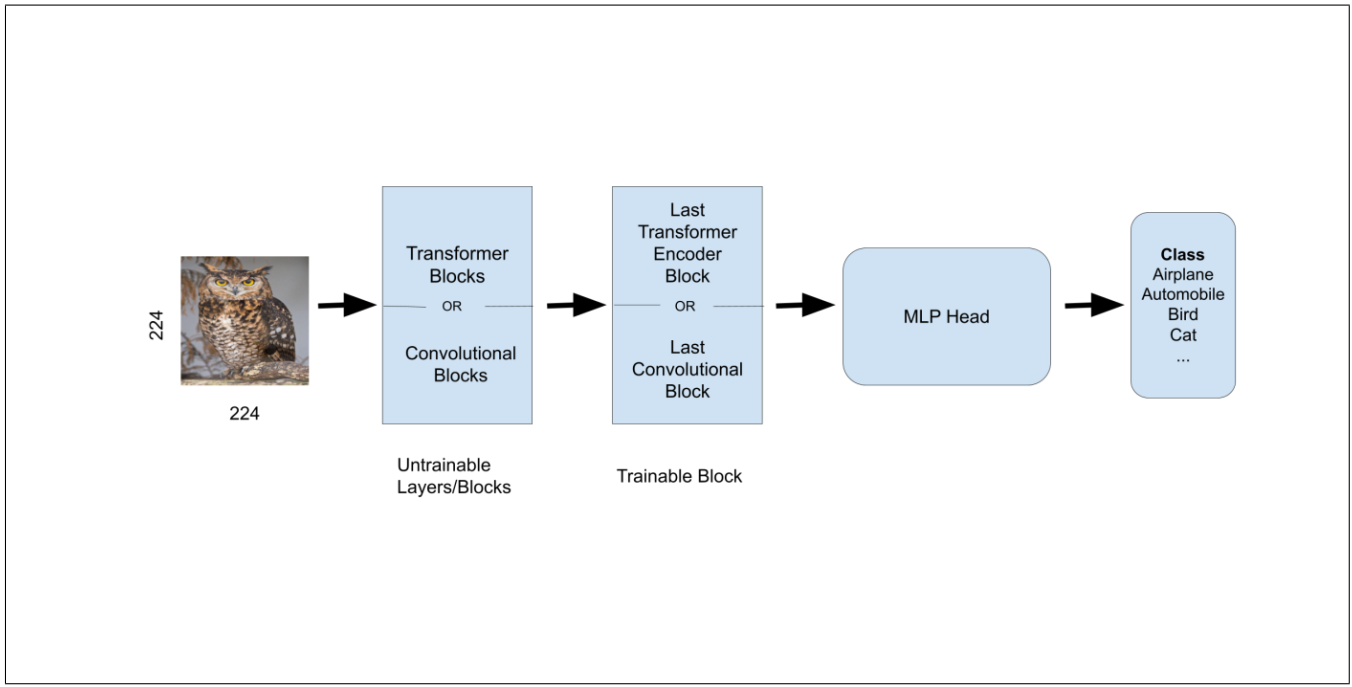

Figure 1. Our general architecture consisting of all the blocks of the Transformers/CNN frozen except for the last block having an MLP head at the end which classifies the image into the output class.

### 3.3. Fine-tune:

When a model training on one task is being re-purposed on some other task then this process is called Transfer Learning. When modelling the second task, transfer learning is an optimization that allows for faster development or better performance. Fine-tuning is closely related with Transfer Learning. It helps us to preserve the weights of previous layers. We freeze all the layers of the model and unfreeze the layers starting from the last block or layer to make the frozen layers non trainable and the last layers trainable. For reproducibility, the block/layer from which we unfreeze for every model is as follows:-

- VGG-19: block5_conv1
- ResNet152V2: conv5_block1_preact_bn
- DenseNet201: conv5_block1_0_bn
- EfficientNetV2-L: Last InvertedResidual of last Sequential block
- NFNetF6: Last NormFreeBlock of last Sequential block
- CoaT-LiteSmall: Last SerialBlock of the serial_blocks_4
- CaiT-S24: Last block of the blocks_token_only module
- DeiTBaseDistilled: 12th Transformer Block

- Swin-B: Last Swim-Transformer block from last BasicLayer

- ViT-L32: Transformer/encoderblock 23

### 3.4. Hyperparameter:

A parameter whose value is determined prior to the start of the learning process is called a hyperparameter. Hyperparameter-tuning is the process of searching for ideal hyperparameters for an architecture. Across all models, we keep the same hyperparameters while fine-tuning to see which architecture learns the best when no special modifications are done to it for training purposes. Learning rate schedules aim to regulate the learning rate during training by lowering it according to a predetermined timetable. We have kept the learning rate to be 0.0001. Number of epochs indicate the number of passes of the dataset that the machine learning algorithm has been completed. A forward and backward pass are counted as a single pass. Since one epoch is too big and hence can't be directly fed to the machine at once, we divide it into mini-batches. We have trained our model on maximum 20 epochs and a batch size of 512 with an exception for Swin-L[15] which had a batch size of 256 due to GPU resource limitations. Too many epochs can lead to overfitting of the model and on the contrary, less number of epochs may lead to under-fitting of the data. Hence we are training all the models for 20 epochs. Also, we reduce the learning rate once validation accuracy stops improving. The learning rate reduces by a factor of 0.6 and has a minimum value of 0.0000001. We have implemented horizontal flip augmentation which is a technique used to artificially expand the size of a dataset by creating modified versions of images by flipping the input on the horizontal axis.

### 3.5. Last layers and Loss:

Flatten, Linear, Dropout, Linear layers in the mentioned order are attached to the end part of the model. This network is also called the head for the Pytorch models. After the last block for all the models which can be either a convolutional layer, convolutional block, transformer encoder block or a transformer encoder, we append these layers to it while keeping the extra layers such as the normalization layers, pooling layers, etc. Our first flatten layer takes in as input the output dimensions of the model and connects it to a set of 256 neurons with an elu[4] activation function. This layer is then regularized with dropout regularization with dropout probability of 0.5%. We then add another layer that connects these 256 neurons to 10 neurons and has a softmax activation. We use a simple Cross Entropy loss to finally classify into the classes of CIFAR-10.

## 4. Results:

In this section, we report and discuss the accuracy obtained by all the models on the validation dataset of CIFAR-10. In Table 1, We report the maximum accuracies reached by every model, their number of trainable parameters, the dataset it was pretrained upon. We plot the accuracy and number of trainable parameters and show the comparison in Figure 2.

We first discuss the results from CNNs. From the 3 classic CNNs we take, 1 of the models, namely ResNet151V2, does not perform comparably with the rest of the models and their accuracy stays well below 90% on the validation set. This model is huge in size but the residual connections help it in avoiding over-fitting in a way that could hurt the validation accuracy. Although training accuracy keeps increasing after a point, validation accuracy plateaus. The remaining 2 classic CNNs, namely, DenseNet201 and VGG19 perform considerably well compared to ResNet151V2. VGG19 reaches a peak of 92.784% while DenseNet reaches 94.757% on the validation set. This makes DenseNet201 a very competitive model with just 7 Million parameters to train for fine-tuning. These results depict that Densenet201 is the only model from the classic CNNs that achieves a really good accuracy.

For the recent CNNs namely NFNetF6 and EfficientNetV2L which report accuracies in the top 50 leader board for ImageNet Classification[1], NFNetF6 achieves an accuracy of 94.35% on the validation set which is only 0.4% less than DenseNet201. But NFNetF6 has 12 Million trainable parameters which is significantly larger than the 7 Million trainable parameters of the DenseNet201. NFNetF6 also has a total number of parameters including trainable and non-trainable parameters, equal to approximately 450 Million[3] which is much larger than DenseNet201 with 18 Million total parameters. This makes the NFNetF6 a worse choice for application in real life due to ultimately more inference time and also a bit less accuracy. As for the EfficientNetV2L, only accuracy of 90.13% is achieved with a number of parameters similar to DenseNet201. From these results, we can understand that even though EfficientNetV2L performs well on the ImageNet dataset, it is not as good at transferring the learned weights to another dataset with our method of fine-tuning. Along with this, DenseNet201 proves to be a better choice for fine-tuning compared to NFNetF6.

Moving on to the transformer family of models, Vision Transformers (ViT) being trained on ImageNet21k, perform quite well achieving an accuracy of 95.6% which is better than DenseNet201 from the CNN family. We don't consider the comparison of this model with any other model as it is pretrained on ImageNet21K. Swin-B reaches an accuracy of 93.58% with 12 Million parameters which is approximately 1% less accurate compared to the DenseNet201. The train-ing for Swin transformers show spikes in the validation accuracy as shown in Figure 3. This makes it hard to understand the number of epochs it is needed to be fine-tuned for. This model performs comparably with the DenseNet201 but still does not surpass it.

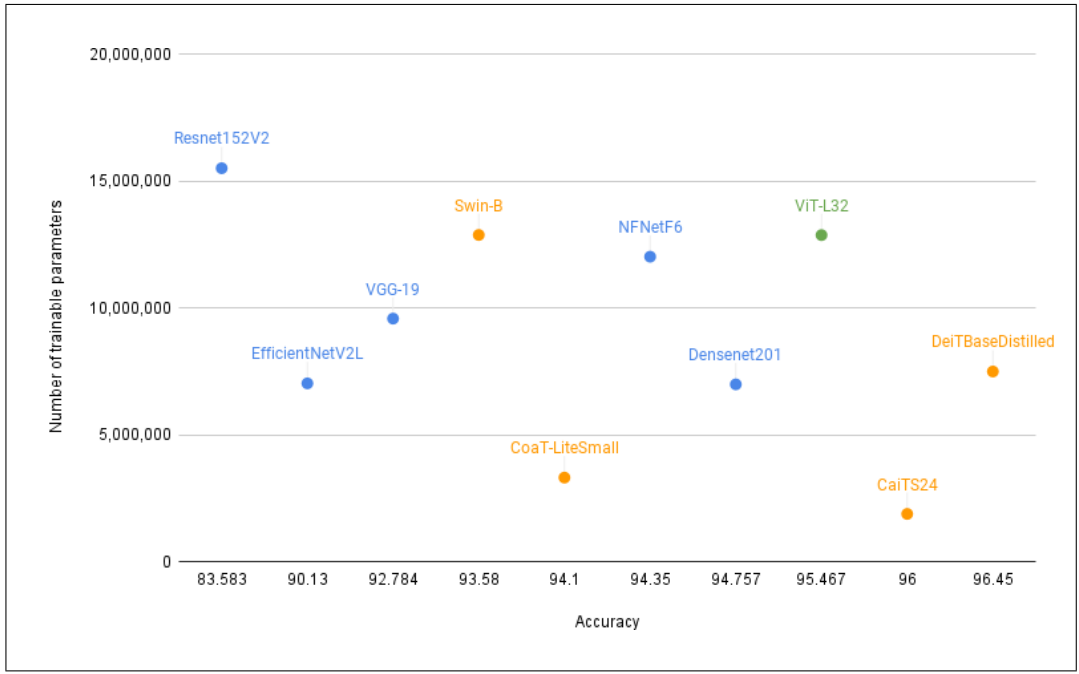

Figure 2. Number of Trainable Parameters vs. Accuracy for all the models including 5 CNNs, 5 Transformers including one Vision Transformer (ViT-L32). A model is considered better than previous one if it has more accuracy or more efficient if it has similar accuracy with a lesser number of trainable parameters.

| Model | Model type | Trainable Parameters | Validation Accuracy(%) | Pretrained on |
|---|---|---|---|---|
| Resnet152V2[9] | CNN | 15,497,994 | 83.583 | ImageNet1K |
| EfficientNetV2L[21] | CNN | 7,020,960 | 90.130 | ImageNet1K |
| VGG-19[17] | CNN | 9,573,130 | 92.784 | ImageNet1K |
| NFNetF6[3] | CNN | 12,012,043 | 94.350 | ImageNet1K |
| Densenet201[10] | CNN | 6,982,400 | 94.757 | ImageNet1K |
| ViT-L32[7] | Transformer | 12,863,242 | 95.467 | ImageNet21K |
| Swin-B224[15] | Transformer | 12,868,650 | 93.580 | ImageNet1K |
| CoaT-LiteSmall[25] | Transformer | 3,308,298 | 94.100 | ImageNet1K |
| CaiTS24[26] | Transformer | **1,877,130** | 96.000 | ImageNet1K |
| DeiTBaseDistilled[22] | Transformer | 7,488,276 | **96.450** | ImageNet1K |

Table 1. Comparison of accuracy obtained on CIFAR-10 validation set by all the models along with their respective trainable parameters and the dataset it was pretrained upon.

The most interesting set of results are achieved by DeiT, CaiT and CoaT models in this experiment. The DeiT-BaseDistilled model achieves the highest 96.45% accuracy with just 7 million parameters. This is approximately 2% better than the DenseNe201's result. The number of parameters are almost similar to DenseNet201. This model outperforms all the previous models including CNNs and Transformers. CoaTLiteSmall which is a smaller model achieves an accuracy of 94.1% which is just 0.65% less than DenseNet201 but only with 3 Million parameters which is half the number of parameters required for DenseNet201.

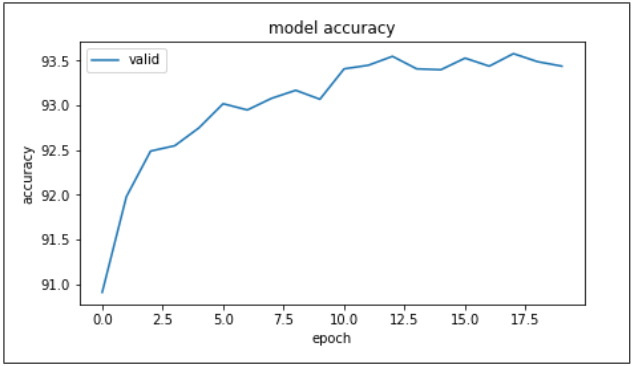

Figure 3. Validation Accuracy vs. epochs while training Swin-B

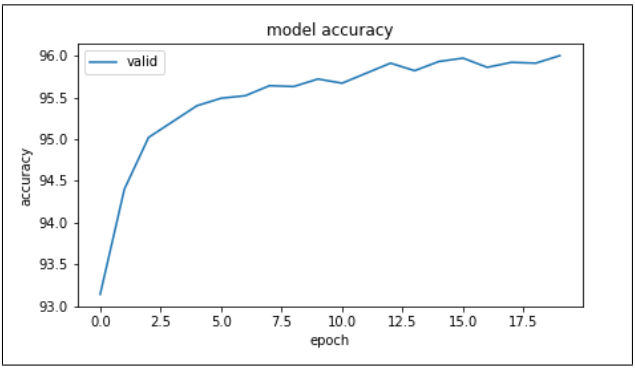

Figure 4. Validation Accuracy vs. epochs while training CaiT

Finally, CaiT achieves an accuracy of 96.0% which is just 0.45% short of the best performing model DeiT. But noticeably, the number of trainable parameters used in this model is only 1.8 Million which is a reduction of four fold in comparison with DeiT. This model also does not show any noticeable spikes in the validation accuracy while training and a smooth curve is observed for this accuracy with respect to epochs as shown in Figure 4. For a balance of efficiency and accuracy, we find out that the CaiT model is best for fine-tuning on CIFAR-10 with our setting. Overall, transformers perform much more efficiently and achieve more accuracy compared to CNNs in the context of transfer learning with limited fine-tuning.

## 5. Conclusion:

We successfully fine-tune all the models on CIFAR-10 with the same simple yet restrictive approach including some classic CNNs, some recent CNNs and recent transformers all of which are pretrained on ImageNet1K and conclude that, while CNNs still achieve high accuracies on ImageNet, the only models to transfer their learning comparably are DenseNet201 and NFNetF6 with accuracies 94.757% and 94.35% respectively while the Transformer based model DeiT achieves the highest accuracy of 96.45% with only 7 Million trainable parameters. The CaiT model which performs 96.0% accuracy which is just 0.45% less than the highest performing DeiT but with only 1.8 Million trainable parameters that is approximately 4 times lesser than the parameters for DeiT. We conclude that for a finetuning task, the models DeiT and CaiT perform extremely well with fewer parameters compared to CNNs. Overall, the transformer based models perform much more efficiently than the CNN based models with the same dataset and finetuning procedures. These results enforce that the transformer architecture is better at transfer learning without any major modifications while fine-tuning and are a better architecture that should be preferred for fine-tuning in the future.